%% file: arxiv_version.tex
\documentclass{article}

\PassOptionsToPackage{numbers,sort&compress}{natbib}
\usepackage{float}
\usepackage{amsmath}
\usepackage{fvextra}
\usepackage[dblblindworkshop, final]{neurips_2026}
\workshoptitle{Sim2Science: ML with Imperfect Scientific Models}
\usepackage[utf8]{inputenc} 
\usepackage[T1]{fontenc}    
\usepackage{hyperref}       
\usepackage{url}            
\usepackage{booktabs}       
\usepackage{amsfonts}       
\usepackage{nicefrac}       
\usepackage{microtype}      
\usepackage{xcolor}         
\usepackage{pifont}
\usepackage{graphicx}
\usepackage{tikz}
\usetikzlibrary{positioning,arrows.meta,fit,calc,backgrounds}
 \usepackage{array}
\usepackage[most]{tcolorbox} 
\usepackage{fancyvrb}
\usepackage{multirow}
\definecolor{promptbar}{HTML}{A6A6A6}   
\definecolor{promptbg}{HTML}{FAFAFA}    
\definecolor{promptrule}{HTML}{C4C4C4}  

\newtcolorbox{promptbox}[2][]{%
  enhanced, breakable,
  colback=promptbg, colframe=promptrule, boxrule=0.6pt,
  arc=4pt, outer arc=4pt,
  title={#2}, fonttitle=\bfseries, coltitle=white, colbacktitle=promptbar,
  titlerule=0pt, toptitle=3pt, bottomtitle=3pt,
  left=8pt, right=8pt, top=7pt, bottom=7pt,
  #1}

\newcommand{\cmark}{\ding{51}}
\newcommand{\xmark}{\ding{55}}
\newcommand{\ocircle}{$\circ$}
\newcommand{\gcheck}{\textcolor{green!60!black}{\cmark}}
\newcommand{\rcross}{\textcolor{red!75!black}{\xmark}}

\title{Solver to System: LLM Reasoning Across Representations and Execution in Scientific Simulation}

\author{%
  \textbf{Eliza Berman\textsuperscript{*}\quad
  Rohan Subramanian\textsuperscript{*}\quad
  Shreya Jha}\\[3pt]
  \textbf{Lohit Jagarapu\quad
  Romain Lopez\quad
  Rajesh Ranganath}\\[6pt]
  New York University\\
  \texttt{\{ehb7466@,rs9768@,sj4964@,laj9771@,rl5696@,rajeshr@cims.\}nyu.edu}
}

\begin{document}

\maketitle

\begingroup
\renewcommand{\thefootnote}{*}
\footnotetext{Equal contribution.}
\endgroup

\begin{abstract}
Simulators based on partial differential equations (PDEs) are widely used to model physical systems, but generating and interpreting PDE solver code with large language models (LLMs) remains challenging. Scientific simulation code is a useful testbed for LLM scientific reasoning because (1) one system can be represented by code, equations, natural language, and numerical trajectories and (2) code can execute successfully while implementing invalid physics. We introduce \emph{Solver2System}, a dataset of PDE solvers with multiple representations, controlled corruptions, and lexical perturbations, to evaluate whether LLMs can recover the underlying physical system through (1) cross-representation consistency and (2) static and agentic detection of physical invalidity. Across eight open-weight models, LLMs systematically favor some representations when evidence conflicts, and lexical perturbations to code alter interpretation of other unchanged representations. Execution improves physical-validity judgments, but weaker agents can judge correctly without identifying the causal defect. Thus, LLMs may reach correct judgments despite inconsistent representation reconciliation or causal misdiagnosis.
\footnote{Code: \href{https://github.com/RohanS14/solver2system}{GitHub};
datasets: \href{https://huggingface.co/datasets/bermaneh/pde-llm-eval-cross-representation-dataset}{cross-representation}
and \href{https://huggingface.co/datasets/bermaneh/pde-llm-eval-code-perturbation-dataset}{code validity} (Hugging Face).}

\end{abstract}

\section{Introduction}

Large language models (LLMs) and AI agents are increasingly used to interact with scientific simulators, including generating, interpreting, and executing numerical partial differential equation (PDE) solvers \citep{IBMResearch, wan2026largelanguagemodelspartial}. These solvers model physical systems across engineering and science, such as heat transfer and fluid flow \citep{LEE20171020, deWit2024}. Developing scientific simulators requires reasoning across different layers of abstraction of the same underlying physical system. We therefore characterize \textbf{scientific simulation code as a representation of that system alongside mathematical equations, natural language, and numerical trajectories}. LLMs are typically evaluated on their ability to use each representation separately, but may underperform in scientific applications that require reasoning across these representations. Reconciling multiple representations of the same underlying system is broadly important for multimodal reasoning, and PDE simulation code offers a controlled instance with precisely specified relationships between representations.

Implementation errors or conflicting specifications can cause different representations of the same physical system to disagree. Judging scientific simulation code is difficult because successful execution does not guarantee physical validity. When LLMs are used to judge whether code is physically valid, they must infer the intended physical system \textit{and} determine if the implementation produces plausible behavior for that system \citep{hang2026pdeagentbenchmultimetricmultilibrarybenchmark, song2026simulationrunssolveswrong}.

In this work, we assess whether LLMs can recover the physical system underlying PDE solvers through two tasks. In the first, \textbf{cross-representation consistency}, we ask whether models can identify disagreements between code, equations, trajectories, and descriptions of the same system, and determine which representation is inconsistent. In the second, \textbf{static and agentic validity judgment}, we ask whether models can assess the physical validity of a solver from its code alone, and how agents use execution to collect evidence and revise their judgments. Across both tasks, we use \textbf{controlled lexical perturbations} to test model sensitivity to code identifiers and comments.

 \paragraph{Contributions.} \textbf{(1)} We introduce the dataset \emph{Solver2System} with aligned PDE solver code, mathematics, language, and numerical trajectory representations, together with valid/invalid variants and lexical code perturbations. \textbf{(2)} We show that models reconcile conflicting representations asymmetrically, and degrading lexical cues in code changes how models resolve inconsistencies in other unchanged representations. \textbf{(3)} We find that execution improves physical-validity judgments and can reduce lexical bias: the best agents combine static reasoning with targeted diagnostics, while weaker agents may find simulation issues through execution without identifying causal code defects.

\section{Related Work}
\label{related_work}

\paragraph{LLMs for generic and scientific code.} LLMs for code generation have been benchmarked on programming and software engineering \citep{chen2021evaluatinglargelanguagemodels,austin2021programsynthesislargelanguage,hendrycks2021measuringcodingchallengecompetence,du2023classevalmanuallycraftedbenchmarkevaluating,jain2024livecodebenchholisticcontaminationfree,jimenez2024swebenchlanguagemodelsresolve}, while comprehension benchmarks test reasoning about existing code behavior \citep{gu2024cruxevalbenchmarkcodereasoning, chen2024reasoningruntimebehaviorprogram, hooda2025largecodemodelsunderstand}. Robustness work (ClassEval-Obf \citep{le2025names}, CodeCrash \citep{Lam2025}) uses obfuscated identifiers and misleading comments to show that lexical cues are reasoning shortcuts for LLMs. Scientific benchmarks, including our work, evaluate LLMs beyond code execution \citep{sun2026scicodepile128gbcorpusexecutable, somasekharan2026cfdllmbenchbenchmarksuiteevaluating, mudur2025feabenchevaluatinglanguagemodels, wan2026largelanguagemodelspartial}. Many studies address the forward problem of generating simulation code given a natural language or mathematical specification \citep{li2025codepde, gaonkar2025scimlagentswritesolver, fazliani2025pdesharppdesolverhybrids, wu2025automatedcodedevelopmentpde, du2026autonumericsautonomouspdeagnosticmultiagent}, and PDEAgent-Bench \citep{hang2026pdeagentbenchmultimetricmultilibrarybenchmark} evaluates LLM-generated solvers using reference simulations. We instead study the problem of \textit{recovering and verifying the physics encoded in existing simulation code}. MooseBench similarly studies where intended physics differs from LLM-generated implementations \citep{song2026simulationrunssolveswrong}. They compare intent specified in natural language with solvers written in structured code frameworks where semantics unambiguously reveal the underlying PDE.  We test deeper physical reasoning by instead asking LLMs to interpret code alongside other unstructured representations,
without reference simulations and under lexical perturbations.

\paragraph{Cross-representation physical reasoning.} Reasoning consistently about a shared physical system across representations is a challenging multimodal evaluation problem for vision language models (VLMs) and vision-language-action (VLA) models \citep{11495231}. Recent VLM work has flagged asymmetric trust \citep{vansprang2026contentdifferentanswerscrossmodal} and systematic biases \citep{pezeshkpour2025mixedsignalsdecodingvlms} across modalities, and proposed benchmarks (PhysBench \citep{chow2025physbenchbenchmarkingenhancingvisionlanguage}) using video-image-text views of a scene. PDE simulation code provides a controlled, more precise setting where representations have known ground-truth relationships and governing laws, enabling direct consistency manipulation to measure which representations LLMs trust.

\textbf{Evaluation of coding agents.} Coding agents allow LLMs to modify and execute code, motivating generic \citep{yang2023intercodestandardizingbenchmarkinginteractive, yang2024sweagentagentcomputerinterfacesenable} and scientific code-focused benchmarks \citep{chen2025scienceagentbenchrigorousassessmentlanguage, duston2025ainsteinbenchbenchmarkingcodingagents} that  evaluate final task success and sometimes turn-by-turn agent behavior. We instead study agentic execution as \textit{scientific evidence acquisition}, asking what evidence an agent collects given an existing simulator, whether execution leads to belief revision, and whether agents can identify causal defects.

\section{Datasets and Evaluation Framework}
\label{sec:evaluation-framework}

Solver2System consists of two datasets derived from a shared corpus of PDE solvers: the \textit{Cross-Representation Dataset}, which tests whether models reconcile code, equations, trajectories, and language describing the same physical system, and the \textit{Code Validity Dataset}, which tests static and agentic judgments of physical validity under lexical perturbations (Figure~\ref{fig:overview}).

\paragraph{Shared Corpus of Simulation Code.}
\label{shared-code-corpus}

We compile 32 implementations of Python PDE solvers spanning Heat, Wave, Burgers, and Navier-Stokes (8 per family, open-source or synthetic), varying physical behavior, dimensionality, numerical scheme, and initial/boundary conditions. We present textbook-style implementations of well-characterized equations rather than high-fidelity code. From each valid solver, we construct a physically invalid implementation by introducing small but consequential defects such as sign flips, spurious terms, and disabled numerical operations. These defects preserve executability while corrupting dynamics, producing failures such as unbounded growth, spurious oscillations, or NaN values. Corpus construction details are in Appendix~\ref{app:dataset-generation}.

\input{figure_tikz_arxiv}

\paragraph{Cross-representation consistency.} For each physical system $s$, consider four aligned views: $R_s = (C_s,T_s,D_s,M_s)$. $C_s$ is PDE solver code, $T_s$ is a numerical trajectory from executing \(C_s\) (subsampled to 10 timepoints), $D_s$ is a natural-language description of the physical process, and $M_s$ is the governing equation with initial/boundary conditions. A \emph{consistent} item has all four views describing the same physical system $s$, while an \emph{inconsistent} item corrupts \textit{exactly one view}. The \textit{Cross-Representation Dataset} contains the four aligned representations $R_s$, with corresponding corruptions for each one, for all $s$. Corrupted code is the physically invalid implementation described above. Corrupted descriptions introduce a small but physically meaningful error. Corrupted mathematics alters an equation term or an initial/boundary condition. We construct four trajectory corruptions: output from the invalid solver, another system's trajectory, within-trajectory permutation, and random values. Models predict whether $R_s$ is consistent and, if not, identify the outlier. We measure disagreement detection, false-alarm rate on consistent items, and false-blame rate by representation. We also evaluate the effect of removing comments and identifier names from source code, which does not affect the consistency and thus tests how sensitive models are to lexical cues.

\paragraph{Static and agentic validity judgment.} The \textit{Code Validity Dataset} contains valid/invalid solver pairs under four lexical conditions: original code, no comments, corrupted comments, and obfuscated identifiers (Appendix~\ref{app:pert_types}). Base solvers contain informative comments and identifiers. For comment corruption, comments are made deliberately misleading by replacing them with those from a different PDE class solver. For identifier obfuscation, comments are absent and variable and function names are replaced with uninformative placeholders. Agents first judge the PDE class, numerical method, physical behavior, and physical validity from static code reading, then may collect evidence through execution before revising the same judgment. We compare pre- and post-execution performance on validity discrimination, then analyze reliance on lexical cues and turn-by-turn actions.

\section{Experiments}

\subsection{Cross-Representation Consistency} \label{sec:cross_representation}

We test how models resolve controlled disagreements across code, mathematics, trajectories, and language. We evaluate eight reasoning-enabled open-weight models (six Qwen-based, one Nemotron, and one GLM), with three sampled generations per item. Models report whether the four views agree, identify the outlier if present, and provide a justification. Model versions, inference settings, prompts, and per-model results are in Appendix~\ref{app:cross_representation}.

\begin{figure}[t]
    \centering
    \includegraphics[width=.95\linewidth]{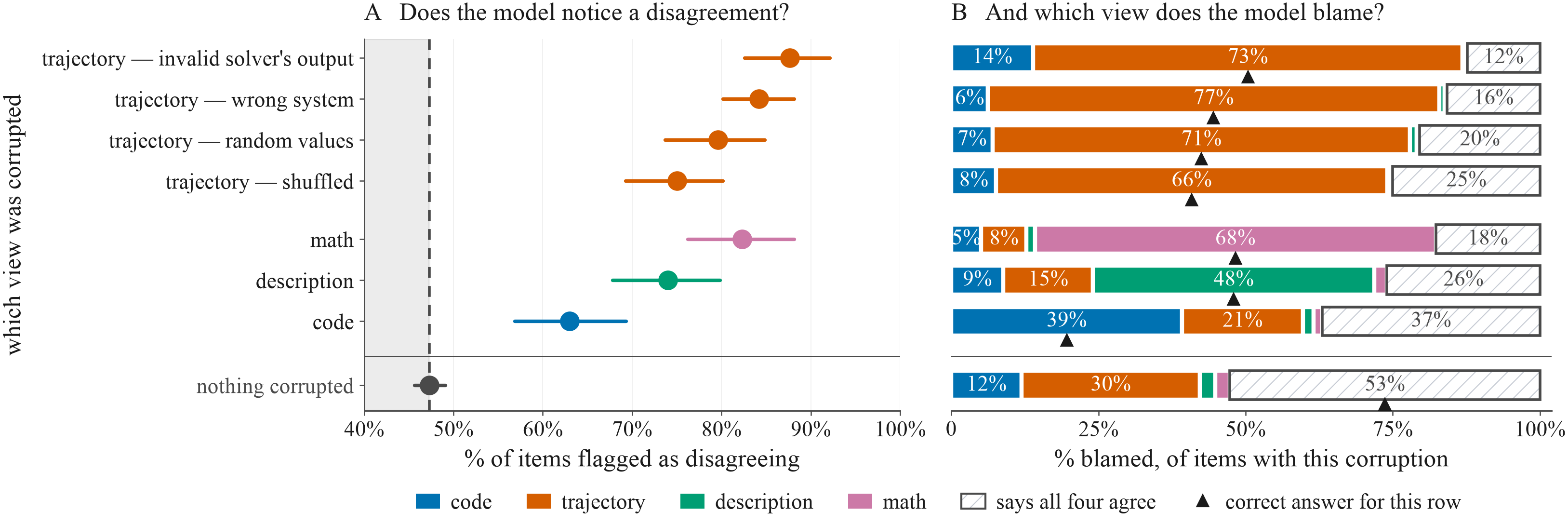}
    \caption{\textbf{A: Disagreement detection by corruption type.} Dashed line marks the 47\% false-alarm rate on consistent items; error bars indicate 95\% CIs. \textbf{B: Inconsistency localization  by corruption type.} For each corruption, bars show which view (or no view) was blamed as the outlier.}
    \label{fig:xr_detection_localization}
\end{figure}

\paragraph{Models are more sensitive to some representation corruptions than others.} Models exhibit a high false alarm rate, incorrectly reporting disagreement on 47\% of consistent items. Nevertheless, corruptions increase the disagreement rate in every condition, from 63\% for corrupted code to 88\% for invalid solver's trajectory (Figure~\ref{fig:xr_detection_localization}A). Corrupted code is least detectable, followed by descriptions and then math. This may reflect memorization of canonical equations, which are easier to pattern-match than stylistically diverse solver code. Trajectory disagreement detection varies from 88\% for execution-based corruptions to 75\% for within-trajectory shuffling. This suggests that models may use structural cues such as numerical blowup in the invalid solver's trajectory or shape differences for swapped trajectories, while within-trajectory shuffling destroys physically meaningful structure while preserving array shape and distributional statistics. Full blame matrix is shown in Appendix~\ref{blame-matrix}.

\paragraph{When evidence conflicts, models exhibit an implicit hierarchy between representations.}
Detecting disagreement alone does not demonstrate cross-representation reasoning: models could always report a disagreement or distrust the same view. We therefore measure how often the model assigns blame to each view and if that view is the true outlier when a disagreement is detected (Figure~\ref{fig:xr_detection_localization}B). Trajectory is the most frequently blamed innocent view across non-trajectory corruptions: blamed on 21\% of items when code is corrupted, 15\% for description corruptions, and 8\% for the math corruptions. In these same conditions, other innocent views receive only 1--9\% of the blame, suggesting that models preferentially distrust numerical trajectories during conflicts.

\begin{figure}[t]
    \centering
    \includegraphics[width=0.95\linewidth]{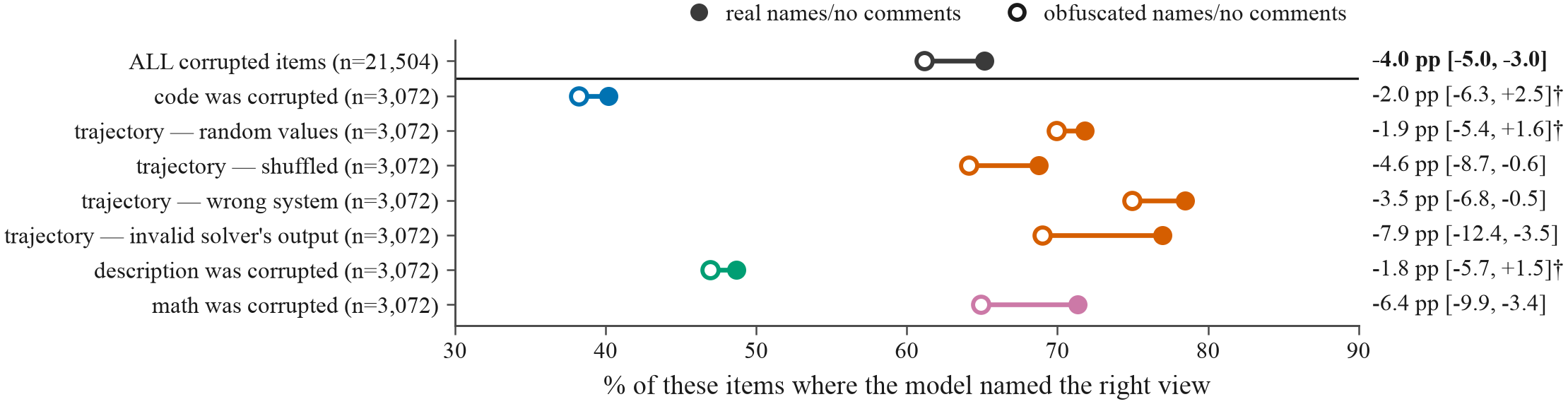}
    \caption{Correct outlier identification with real ($\bullet$) vs.\ obfuscated ($\circ$) identifiers, both without comments. $\Delta =$ obfuscated $-$ real (pp); 95\% Bonferroni-corrected solver-clustered paired-bootstrap CIs (32 solvers). $\dagger$ denotes confidence intervals containing zero.}
    \label{fig:xr_obfuscation}
\end{figure}

\paragraph{Obfuscating identifiers in source code makes cross-representation inconsistencies less detectable.}

Finally, we test whether code identifiers change how models reconcile conflicting evidence. We compare informative and obfuscated identifiers, removing comments in both conditions. Obfuscation reduces correct outlier identification by 4.0pp across corrupted items (95\% CI $[-5.0, -3.0]$). Notably, this effect is not isolated to cases where the code is corrupted. The largest drops occur for \emph{invalid solver's output trajectory} (-7.9pp) and corrupted \emph{math} (-6.4pp) (Figure~\ref{fig:xr_obfuscation}), even though these representations are unchanged. Intervals  exclude zero for shuffled and wrong-system trajectories, but include zero for code, random-value trajectories, and descriptions. Interestingly, in these cases, the model increasingly flags all items as in agreement, suggesting that identifier obfuscation makes cross-representation inconsistencies less detectable.

\subsection{Static and Agentic Validity Judgments} \label{sec:agents}

We ask how models revise static judgments of physical validity using evidence from agentic execution. We find that static judgments by open-weight models are affected by lexical cues in code (Appendix~\ref{app:validity-judgments}). 
Here, we evaluate two frontier models, Gemini 2.5 Flash \citep{google2025gemini25flash} and 3.7 Flash \citep{google2026gemini3.7flash}, under two thinking settings. 
Models first judge validity (yes/no \textit{and} 0–10 score) statically, then may revise judgments after at most six actions that (1) modify the simulator to generate new evidence or (2) run a diagnostic to analyze existing evidence. Prompts, tools, and harness details are in Appendix~\ref{app:agentic}. 

\begin{figure}[t]
    \centering
\includegraphics[width=.95\linewidth]{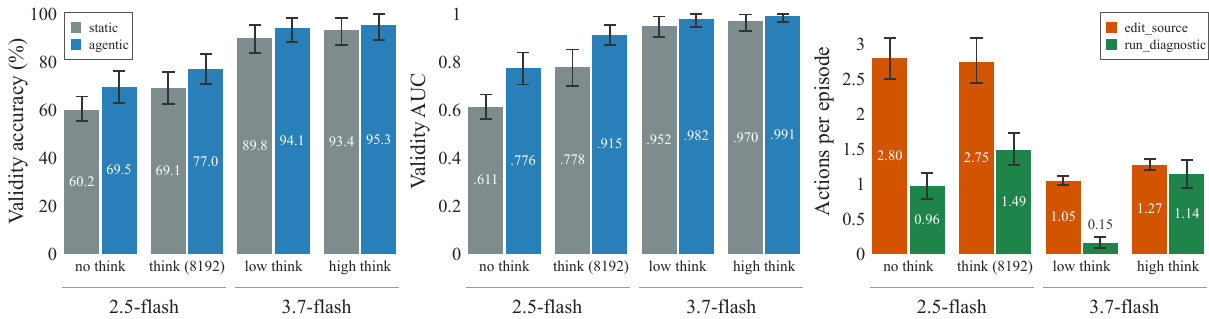}
    \caption{Validity accuracy, score AUC, and  action breakdown for different agents. Error bars show 95\% cluster-bootstrap confidence intervals (effective $n=32$ independent ground truth scripts).}
    \vspace{-5pt}
    \label{fig:agentic1}
\end{figure}

\paragraph{Execution improves validity judgments and stronger agents use fewer actions.} Across models and thinking conditions, execution improves physical-validity judgments (Figure~\ref{fig:agentic1}). Improvements are largest when static performance is weakest: Gemini 2.5 without thinking gains 9.3pp accuracy after execution, while Gemini 3.7’s strong static reasoning (93\% static accuracy with high thinking) is near-ceiling and gains just 1.9pp. Gemini 3.7 also uses fewer actions than Gemini 2.5 (2.42 vs. 4.24 at high thinking), and prefers targeted diagnostics to repeatedly modifying the simulator--although diagnostics may miss physically inconsistent behavior (Gemini 3.7 example in Appendix~\ref{qual-failure-case}).

\begin{figure}[t]
    \centering
    \includegraphics[width=0.95\linewidth]{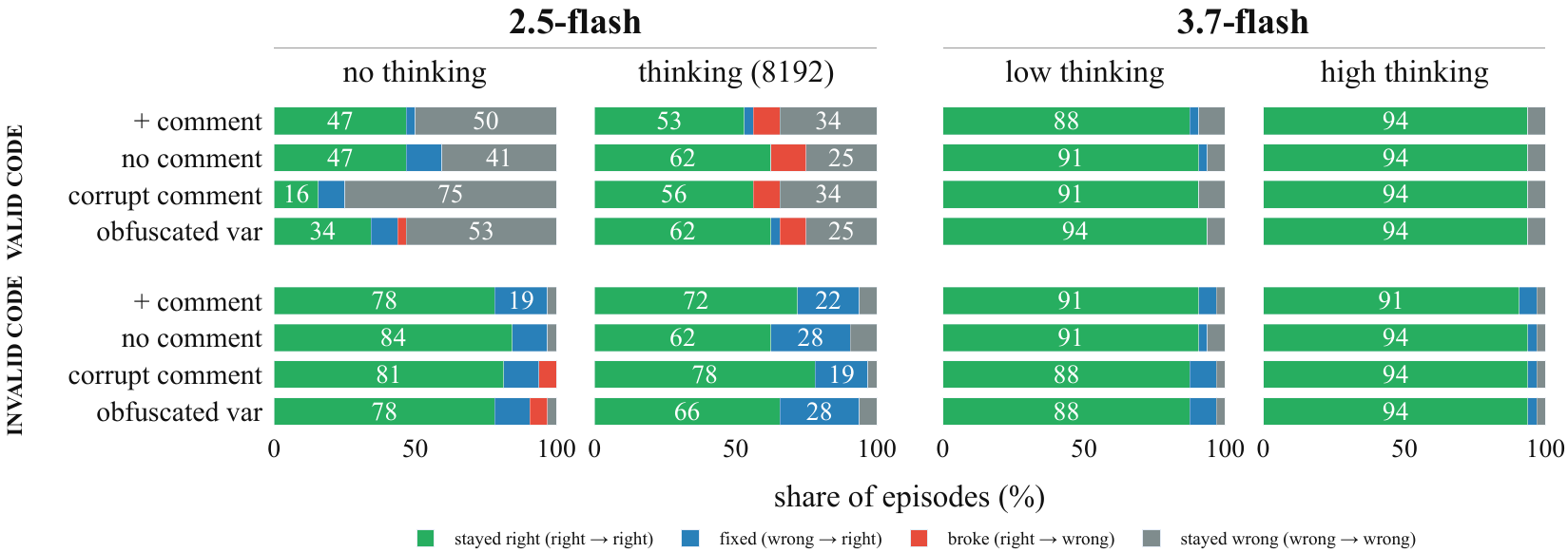}
    \caption{Evaluating how agents revise validity judgments across lexical perturbations to code.}
    \label{fig:agentic2}
    \vspace{-5pt}
\end{figure}

\paragraph{Execution can reduce lexical sensitivity, but may mislead weaker models.} Sensitivity to lexical cues varies across models and thinking settings (Figure~\ref{fig:agentic2}).
Under corrupted comments, recall on valid snippets for Gemini 2.5 (no thinking) is just 25\% after execution and some correct invalid-calls are reversed. Inversions (an invalid solver receives a higher validity score than its valid counterpart) move from 12\% to 13\% after execution, suggesting that execution was misleading. Gemini 2.5 with thinking benefits more: execution reduces inversions from 15\% to 6\% (Appendix~\ref{app:additional-validity-agents-results}). Gemini 3.7 is robust to lexical cues, correcting judgments at low thinking mainly when comments or variables are perturbed and never reversing correct judgments. Differences between Gemini versions, however, prevent attributing this robustness to model reasoning capability alone.

\begin{figure}[h]
    \centering
    \includegraphics[width=0.95\linewidth]{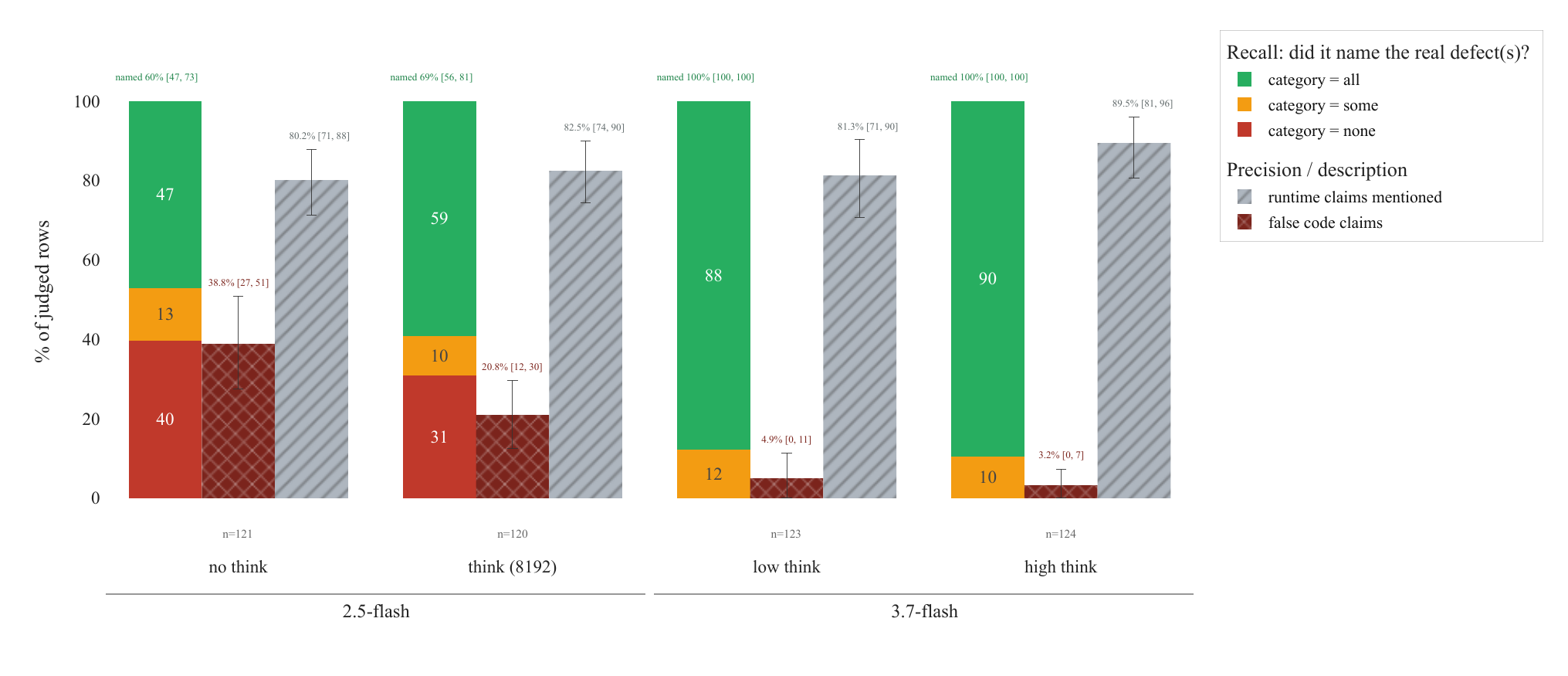}
    \caption{Evaluating reasoning quality for each agent using LLM-as-judge.}
    \label{fig:agenticsupp2}
\end{figure}

\paragraph{Correct validity judgments can still rest on incorrect causal reasoning.} We ask whether agents that identify invalid solvers also find the responsible defects. For correctly identified invalid snippets, we evaluate the model’s justification with LLM-as-a-judge (see Appendix~\ref{llm-as-judge}), asking whether the model makes an incorrect claim about the code, and whether it misses any causal defects. We find that Gemini 2.5 frequently reaches the right answer for the wrong reason (21\% of justifications make false code claims and 31\% miss all causal defects at high thinking). Sign flips are often missed while their numerical effects are observed and misattributed to other parts of the code (Appendix~\ref{qual-failure-case}). Contrastingly, Gemini 3.7 identifies at least one genuine corrupted line of code in every case, and false code claims occur in only 4.9\% of low thinking and 3.2\% of high thinking cases (Figure~\ref{fig:agenticsupp2}). 

\section{Conclusion}
\label{sec:conclusion}

We introduce \emph{Solver2System}, a dataset of PDE solvers with aligned code, equation, trajectory, and text representations, to evaluate whether LLMs can recover and verify an underlying physical system. LLMs may reconcile conflicting evidence asymmetrically across representations, and agents may correctly judge a simulator’s physical validity without causal diagnosis. Future work can relate performance on representation consistency to validity judgment by evaluating a shared set of frontier models on both tasks. The most recent models evaluated here show high but not perfect performance, so an expanded benchmark on frontier models could reveal the extent to which the issues identified persist. We position cross-representation reconciliation and validity judgment as useful tests for LLM reliability across a broader array of scientific tasks.

\bibliography{refs-clean}


\clearpage
\appendix

\section{Dataset generation}
\label{app:dataset-generation}

\subsection{Dataset summary statistics by PDE class}
\label{dataset-summary-stats}

\begin{table}[H]
    \centering
    \resizebox{\textwidth}{!}{%
    \begin{tabular}{llrrrll}
    \toprule
    \textbf{Source} & \textbf{PDE} & \textbf{Snippets} & \textbf{Avg Lines} & \textbf{Avg Chars} & \textbf{Methods Covered} & \textbf{Behaviors Covered} \\
    \midrule
    \multirow{5}{*}{Human}
      & Burgers       & 4 & \phantom{00}67 (50--80)  & 1{,}655 (1{,}332--2{,}102) & Explicit                     & Advection \\
      & Heat          & 4 & \phantom{00}40 (36--43)  & \phantom{0,}742 (631--784)   & Explicit, Implicit           & Diffusion \\
      & Navier-Stokes & 4 & \phantom{0}128 (95--177) & 4{,}879 (3{,}356--6{,}459) & Explicit, Implicit, Spectral & Advection, Diffusion \\
      & Wave          & 4 & \phantom{00}66 (44--93)  & 3{,}230 (998--7{,}666)      & Explicit, Implicit, Spectral & Oscillation, Restoration \\
    \cmidrule{2-7}
      & \textbf{Subtotal} & \textbf{16} & \phantom{00}75 (36--177) & 2{,}626 (631--7{,}666) & & \\
    \midrule
    \multirow{5}{*}{Synthetic}
      & Burgers       & 4 & \phantom{00}36 (29--43)  & \phantom{0,}983 (763--1{,}312) & Explicit                     & Advection, Diffusion \\
      & Heat          & 4 & \phantom{00}38 (29--51)  & \phantom{0,}886 (670--1{,}319) & Explicit, Implicit, Spectral & Diffusion \\
      & Navier-Stokes & 4 & \phantom{00}51 (32--75)  & 1{,}490 (749--2{,}709)      & Explicit, Implicit, Spectral & Advection, Diffusion, Restoration \\
      & Wave          & 4 & \phantom{00}42 (33--50)  & 1{,}173 (874--1{,}435)      & Explicit, Implicit, Spectral & Oscillation, Restoration \\
    \cmidrule{2-7}
      & \textbf{Subtotal} & \textbf{16} & \phantom{00}42 (29--75) & 1{,}133 (670--2{,}709) & & \\
    \midrule
    \multicolumn{2}{l}{\textbf{Total}} & \textbf{32} & \phantom{00}58 & 1{,}880 & & \\
    \bottomrule
    \end{tabular}}
    \caption{Ground truth commented dataset statistics by PDE type and source. Each base snippet is replicated across 8 conditions, yielding 64 rows per PDE type (256 total).}
    \label{tab:dataset}
\end{table}

\subsection{Sources of PDE solvers}
\label{app:pde-solvers-sources}

We reduce the open-source implementations to self-contained Python scripts containing the core numerical simulation logic. Sources of human-generated PDE solvers are shown in Table~\ref{pde-sources}. Solvers are redistributed under their original licenses. All other code snippets were generated with Claude Code as described in Appendix~\ref{app:data-gen-prompts}.

\begin{table}[H]
\centering
\label{tab:pde_solver_sources}
\renewcommand{\arraystretch}{1.15}
\setlength{\tabcolsep}{6pt}
\begin{tabular}{p{0.13\linewidth} p{0.15\linewidth} p{0.60\linewidth}}
\toprule
\textbf{PDE} & \textbf{Code Snippet} & \textbf{Source} \\
\midrule

\multirow{4}{0.13\linewidth}{Heat Equation}
& heat\_1 & \href{https://github.com/gilbertfrancois/partial-differential-equations/tree/master/notebook}{Gilbert Francois - Heat Equations} \\
& heat\_2 & \href{https://github.com/gilbertfrancois/partial-differential-equations/tree/master/notebook}{Gilbert Francois - Heat Equations} \\
& heat\_3 & \href{https://github.com/gilbertfrancois/partial-differential-equations/tree/master/notebook}{Gilbert Francois - Heat Equations} \\
& heat\_4 & \href{https://github.com/gilbertfrancois/partial-differential-equations/tree/master/notebook}{Gilbert Francois - Heat Equations} \\
\midrule

\multirow{4}{0.13\linewidth}{Wave Equation}
& wave\_1 & Wave Equation Simulations \cite{binder2021shallowwater} \\
& wave\_2 & Wave Equation Simulations \cite{binder2021shallowwater} \\
& wave\_3 & \href{https://github.com/gilbertfrancois/partial-differential-equations/tree/master/notebook}{Gilbert Francois - Wave Equations} \\
& wave\_4 & Wave Equation Simulations \cite{binder2021shallowwater}, with LLM-assisted modifications to increase numerics diversity\\
\midrule

\multirow{4}{0.13\linewidth}{Burgers Equation}
& burgers\_1 & \href{https://github.com/gilbertfrancois/partial-differential-equations/tree/master/notebook}{Gilbert Francois - Burgers Equations} \\
& burgers\_2 & \href{https://github.com/gilbertfrancois/partial-differential-equations/tree/master/notebook}{Gilbert Francois - Burgers Equations} \\
& burgers\_3 & \href{https://github.com/gilbertfrancois/partial-differential-equations/tree/master/notebook}{Gilbert Francois - Burgers Equations} \\
& burgers\_4 & \href{https://github.com/gilbertfrancois/partial-differential-equations/tree/master/notebook}{Gilbert Francois - Burgers Equations} \\
\midrule

\multirow{4}{0.13\linewidth}{Navier Stokes Equation}
& navier\_stokes\_1 & CFD Python  \cite{barba2018cfdpython} \\
& navier\_stokes\_2 & CFD Python  \cite{barba2018cfdpython}  \\
& navier\_stokes\_3 & PDEBench \cite{takamoto2022pdebench} \\
& navier\_stokes\_4 & Jax CFD \cite{kochkov2021mlaccelerated} \\

\bottomrule
\end{tabular}
\caption{Sources of PDE solver implementations used in experiments}
\label{pde-sources}
\end{table}

\subsection{Synthetic data generation prompts}
\label{app:data-gen-prompts}

Below we list the four prompts used to synthetically generate code implementations for each PDE class via Claude Code \citep{claudecode}, as described in Section~\ref{sec:evaluation-framework}.

\begin{promptbox}{Heat Equation Prompt}
\begin{Verbatim}[fontsize=\small]
Generate 4 fully runnable, end-to-end Python scripts for solving the
heat equation (du/dt = alpha * grad^2 u). Diversity Requirements:

* Spatial Dimensions: Mix 1D and higher dimensions (e.g., two 1D
  scripts, two 2D/3D scripts).
* Numerical Schemes: Mix different numerical schemes (e.g., FTCS,
  Crank-Nicolson, ADI, Fourier Spectral, or Implicit Euler)
* Initial Conditions: Use distinct initial condition functions (e.g.,
  Gaussian distribution, top-hat/step function, sinusoidal wave, and
  a multi-point heat source).

Code Formatting Rules:

* Each script must be complete, runnable, and save or display a plot
  of the solution.
* Keep comments minimal: strictly 5 to 6 inline/docstring comments
  per script.
* Ensure all stability conditions (like the CFL condition for
  explicit schemes) are satisfied so the simulation produces valid
  physical results.
\end{Verbatim}
\end{promptbox}

\begin{promptbox}{Wave Equation Prompt}
\begin{Verbatim}[fontsize=\small]
Generate 4 fully runnable, end-to-end Python scripts for solving the
wave equation (d^2u/dt^2 = c^2 * grad^2 u). Diversity Requirements:

* Spatial Dimensions: Include a mix of 1D and 2D spatial setups
  (e.g., two 1D scripts, two 2D scripts).
* Numerical Schemes: Use different numerical approaches (e.g.,
  standard FDTD Leapfrog, Pseudo-Spectral FFT, Newmark-beta or
  implicit scheme, and finite difference with varying speed/
  absorbers).
* Initial Conditions: Mix initial displacements u(x,0) and initial
  velocities du/dt(x,0) across the scripts

Code Formatting Rules:

* Each script must be complete, runnable, and save/display a plot or
  animation of the wave propagation.
* Keep comments minimal: strictly 5 to 6 inline/docstring comments
  per script.
* Ensure all stability parameters meet the CFL condition
\end{Verbatim}
\end{promptbox}

\begin{promptbox}{Burgers' Equation Prompt}
\begin{Verbatim}[fontsize=\small]
Generate 4 fully runnable, end-to-end Python scripts for solving
Burgers' Equation (du/dt + u * grad(u) = nu * grad^2 u). Diversity
Requirements:

* Spatial Dimensions: Mix 1D and 2D spatial domains (e.g., three 1D
  variations, one 2D vector system).
* Viscosity Regimes: Include both Inviscid (nu = 0) shock-capturing
  behavior and Viscous (nu > 0) smoothing behavior.
* Numerical Schemes: Use a mix of numerical schemes (e.g., Godunov/
  Upwind, Lax-Wendroff, Finite Differences, etc).
* Initial Conditions: Mix continuous waves (sinusoidal), discontinuous
  profiles (shocks/top-hat), and smooth localized pulses (Gaussian).

Code Formatting Rules:

* Each script must be complete, runnable, and save or display a plot
  of the velocity profile.
* Keep comments minimal: strictly 5 to 6 inline/docstring comments
  per script.
* Ensure all stability parameters respect both advective CFL and
  diffusion stability bounds.
\end{Verbatim}
\end{promptbox}

\begin{promptbox}{Navier Stokes Equation Prompt}
\begin{Verbatim}[fontsize=\small]
Generate 4 fully runnable, end-to-end Python scripts for solving the
incompressible/compressible Navier-Stokes Equations:

  du/dt + (u . grad)u = -(1/rho) grad(p) + nu * grad^2 u + f,
  div(u) = 0

Diversity Requirements:

* Spatial Dimensions & Models: Mix 1D and 2D spatial domains
* Viscosity & Flow Regimes: Include high Reynolds number /
  near-inviscid flow (nu -> 0 or Euler equations) highlighting
  advective/shock features, and moderate-to-low Reynolds number
  viscous flow (nu > 0) showing boundary layer or diffusion
  behavior.
* Numerical Schemes: Use a mix of numerical schemes (e.g., Projection
  Method, Finite Differences, spectral methods, etc).
* Initial / Boundary Conditions: Mix different flow configurations

Code Formatting Rules:

* Each script must be complete, self-contained, fully runnable
  (using standard libraries like numpy, scipy, matplotlib), and save
  or display a plot of the results (e.g., velocity field, pressure
  field, streamline contour, or vorticity map).
* Keep comments minimal: strictly 5 to 6 inline/docstring comments
  per script.
* Ensure all time-stepping parameters strictly respect both
  advective CFL and viscous diffusion stability bounds.
\end{Verbatim}
\end{promptbox}

\subsection{Details on code perturbations}
\label{app:pert_types}

Dataset perturbations are shown in Table~\ref{tab:pert_types}. We consider three perturbation operations applied to both physically valid and invalid code:
\begin{enumerate}
    \item \textbf{Comment removal:} all comments are removed from the solver, eliminating natural-language explanations while leaving executable code unchanged.
    \item \textbf{Comment corruption:} comments are replaced with comments drawn from a solver for a different PDE class and numerical method. To corrupt a given snippet (the ``receiver''), we select a ``donor'' snippet at random from valid, commented snippets from 1) a different PDE class, 2) a different numerical method, and (3) the same source code type (human-written vs synthetic). The same donor is used for valid and invalid variants of the receiver. Next, donor comments are extracted in order and inserted exactly at the receiver's comment positions. If the donor has fewer comments than the receiver, comments are cycled, and extra donor comments are discarded. All dataset perturbations were performed using rule-based scripting rather than LLM automation to ensure reproducibility.
    \item \textbf{Variable/function obfuscation:} variable names are replaced with uninformative identifiers (\texttt{foobar1}, \texttt{foobar2}, etc.), and function names are replaced with identifiers such as \texttt{function1}. The renaming is consistent within each snippet, so logic is conserved. 
\end{enumerate}

\begin{table}[H]
\centering
\setlength{\tabcolsep}{4pt}
\begin{tabular}{lccc}
\toprule
\textbf{Perturbation} & \textbf{Comments} & \textbf{Variables} & \textbf{Valid} \\
\midrule
Clean+Comment             & \gcheck & \cmark & \cmark \\
Clean, No Comment         & \ocircle & \cmark & \cmark \\
Corrupt Comment                 & \rcross & \cmark & \cmark \\
Obf Variable         & \ocircle & \xmark & \cmark \\
Invalid+Comment           & \gcheck & \cmark & \xmark \\
Invalid, No Comment         & \ocircle & \cmark & \xmark \\
CorrComment+Invalid     & \rcross & \cmark & \xmark \\
ObfVar+Invalid & \ocircle & \xmark & \xmark \\
\bottomrule
\end{tabular}
\vspace{5pt}
\caption{Dataset perturbations. In the Comments column, \gcheck  \ denotes informative comments, \rcross \ denotes corrupted comments, and \ocircle denotes no comments. In the Variables and Valid columns, \cmark/\xmark \ denote ground-truth vs.\ obfuscated variables and valid vs.\ invalid code, respectively.}
\label{tab:pert_types}
\end{table}

We combine these perturbations into eight conditions, shown in Table~\ref{tab:pert_types}. Each base solver appears once in each condition, producing 256 total snippets: 32 base solvers $\times$ 8 variants.

\section{Cross-representation consistency: models and prompt formats}
\label{app:cross_representation}

\subsection{Models}
\label{app:xr_models}

Table~\ref{tab:xr_models} presents the corresponding model versions for the LLMs
used in our cross-representation experiments. All models are open-weight and were
imported from HuggingFace and served locally.

\begin{table}[h]
\centering
\small
\begin{tabular}{@{}ll@{}}
\toprule
LLM & Model Version  \\
\midrule
R1-Distill-32B      & \texttt{deepseek-ai/DeepSeek-R1-Distill-Qwen-32B}    \\
QwQ-32B             & \texttt{Qwen/QwQ-32B}                                \\
Qwen3-32B           & \texttt{Qwen/Qwen3-32B}                               \\
Nemotron-3-Nano-30B & \texttt{nvidia/NVIDIA-Nemotron-3-Nano-30B-A3B-BF16}  \\
GLM-4.7-Flash       & \texttt{zai-org/GLM-4.7-Flash}                        \\
Qwen3.5-27B         & \texttt{Qwen/Qwen3.5-27B}                            \\
Qwen3.6-27B         & \texttt{Qwen/Qwen3.6-27B}                             \\
Qwen3.8-27B         & \texttt{Qwen/Qwen3.8-27B}                             \\
\bottomrule
\end{tabular}
\caption{LLMs and model versions used in the experiments.}
\label{tab:xr_models}
\end{table}

\subsection{Inference configuration}
\label{app:xr_inference}
We use the following sampling parameters for all models: $T = 0.6$, $p = 0.95$, top-$k = 20$. Every model is run reasoning-enabled. For each prompt, we sample $k=3$ generations per model. The context window is set to each model's full available context, capped at $69{,}632$ tokens. We use sampling rather than
greedy decoding because the reasoning-model cards state that greedy decoding
induces repetition.

\subsection{Compute resources}
\label{app:xr-compute-resources}

All open-weight inference was run on an internal SLURM cluster, on a single NVIDIA H200 GPU per job with 8 CPU cores, 128\,GB of host RAM. Models were downloaded from HuggingFace and served locally with vLLM. All compute is inference only.

This experiment consumed 355.6M prompt tokens and 216.0M generated tokens in total; prompts are long (median 8{,}982 tokens, maximum 32{,}527) because rendered trajectory tables for 2-D fields dominate them. 

\subsection{Inference prompt}
\label{app:xr_prompt}

The following prompt was used:

\begin{promptbox}{Cross-Representation Experiment Prompt}
\begin{Verbatim}[fontsize=\small]
Below are four representations of a physical system, labelled View 1 to View 4.

{views}

Do these representations all describe the same physical system? Either all four
representations are in agreement, or exactly one of them is inconsistent with the
other three.

Answer with a JSON object containing exactly these fields:
- "agree": "yes" if all four agree, "no" if one is inconsistent.
- "outlier": which view is inconsistent, as "view_1", "view_2", "view_3" or
  "view_4". Use "none" if and only if "agree" is "yes".
- "system_pde_class": the class of partial differential equation that the majority
  of the views describe.
- "system_num_method": the numerical method that the majority of the views use.
- "justification": what specifically is inconsistent, or why you judge the views
  consistent.

Output only the JSON object.
\end{Verbatim}
\end{promptbox}

The views are fixed strings, one per representation:

\begin{verbatim}
Python source code for a numerical solver
the governing equation, in LaTeX
the numerical solution field produced by running a solver
a natural-language description of the physical process
\end{verbatim}

Trajectories are rendered with a fixed header stating what the axes mean and
explicitly disclaiming units. The stored trajectory is a bare array with no
accompanying axis metadata (coordinates, time step, grid spacing).

\begin{promptbox}{Trajectory Prompt Header}
\begin{Verbatim}[fontsize=\small]
Numerical solution field, sampled on a uniform normalized grid.
Rows are successive time samples from the start of the run (t=0.00) to the end
(t=1.00) in equal steps.
Within a row, values run over the spatial domain from 0.00 to 1.00 in equal steps
along each axis.
No physical units, grid spacing, or time step are available for this field.

t=0.00:  +3.2500e+02  +3.2500e+02  +3.2500e+02  ...
\end{Verbatim}
\end{promptbox}

\subsection{Experimental design and item count}
\label{app:xr_design}

 Every item presents the same four views of one solver system: code, governing equations, natural-language description, and solution trajectory. Exactly one view is corrupted, or none. Each evaluation item is then defined by four factors: \texttt{SYSTEM|CONDITION|IDENTIFIERS|ORDER}. Table~\ref{tab:xr_design}
summarizes the levels of each factor.

\begin{table}[h]
\centering
\small
\begin{tabular}{@{}lcl@{}}
\toprule
Factor & Levels & Description \\
\midrule
Solver system & 32 &
4 PDE families: Burgers, Heat, Navier--Stokes, and Wave (8 each) \\
Corruption condition & 8 &
1 consistent condition and 7 corrupted conditions, including 4 trajectory corruptions \\
Identifier regime & 2 &
informative identifiers or semantically meaningless identifiers \\
Presentation order & 2 &
two orderings of the four views per system--condition--identifier combination \\
\bottomrule
\end{tabular}
\caption{Factors defining the cross-representation consistency experiment.}
\label{tab:xr_design}
\end{table}

The full factorial design therefore contains $32 \times 8 \times 2 \times 2 = 1{,}024$
items per model. Each item is sampled three times, producing $3{,}072$ generations
per model.

Slot order is permuted
per item, with the corrupted view counterbalanced across slots, so that position
carries no information about which view is the outlier. Each configuration of four views is shown twice, in two independently shuffled orders.

For Figure~\ref{fig:xr_obfuscation}, we estimate paired accuracy differences between identifier conditions and bootstrap the 32 base solvers, retaining all observations from each sampled solver under both conditions. Confidence intervals are Bonferroni-adjusted across eight comparisons.

\subsection{Blame matrix across corruption types}
\label{blame-matrix}

\begin{figure}[H]
    \centering
    \includegraphics[width=0.75\linewidth]{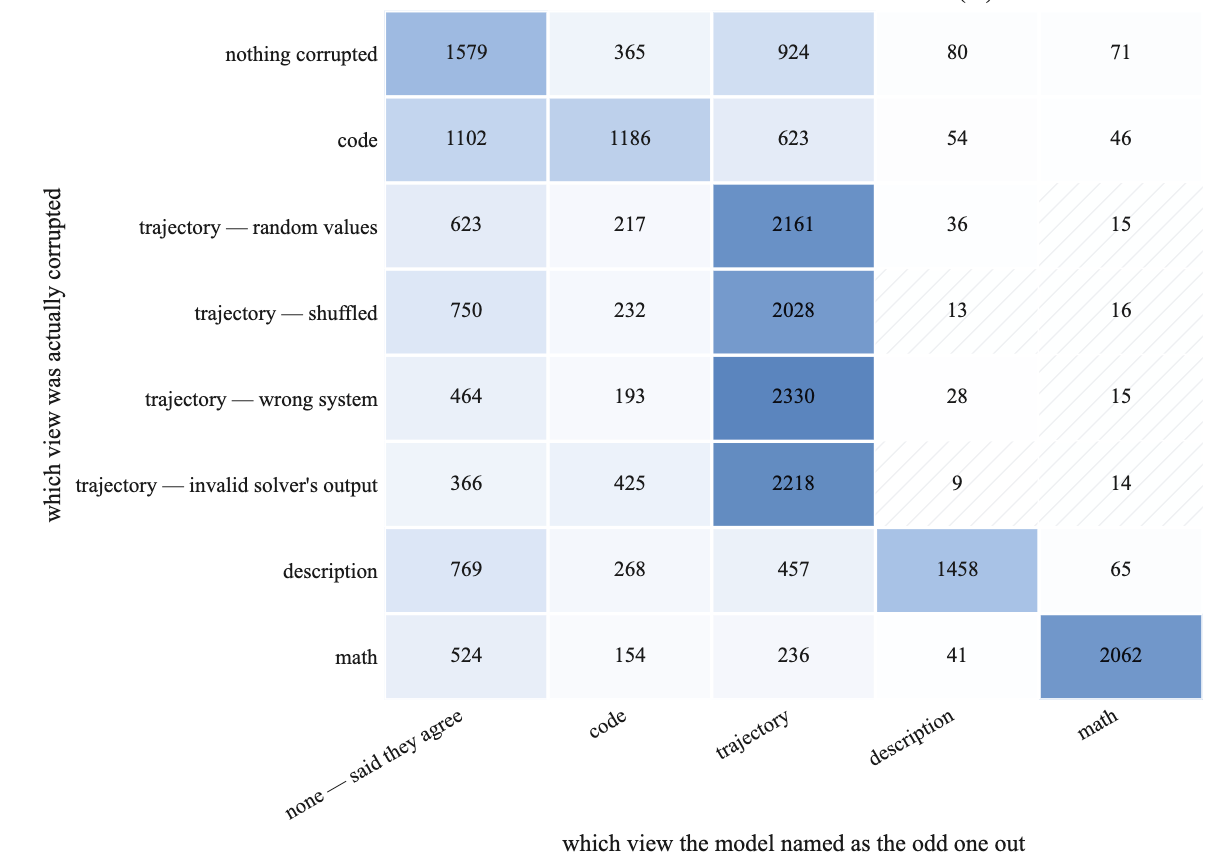}
    \caption{Blame matrix across corruption types. Rows indicate the representation actually corrupted and columns indicate the
view named as the outlier. Diagonal entries are correct localizations;
off-diagonal entries show misattribution.}
    \label{fig:xr_blame}
\end{figure}

Figure~\ref{fig:xr_blame} shows the full confusion pattern between the representation that was actually corrupted and the representation named by the
model as the outlier. The diagonal corresponds to correct localization, while off-diagonal entries reveal systematic misattribution. In particular, when
code, description, or mathematics is corrupted, trajectories receive more incorrect blame than any other innocent representation. The matrix also reflects the substantial rate at which models fail to detect an
inconsistency.

\subsection{Localization accuracy and false-alarm rate across Qwen model family}
\label{app:qwen-ladder}

\begin{figure}[H]
    \centering
    \includegraphics[width=\linewidth]{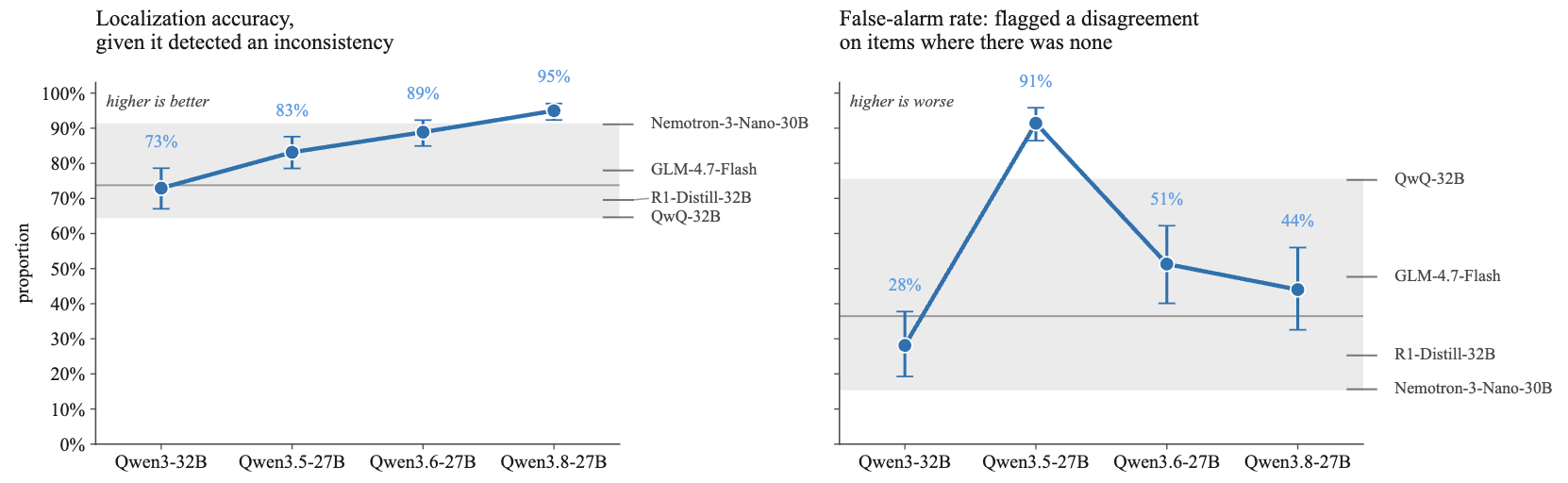}
    \caption{Localization accuracy conditioned on the model having detected an inconsistency (left). On the items where all four views agreed, how often the model flagged an inconsistency anyway (right). While localization accuracy improves with the newer Qwen models, false-alarm rate does not.}
    \label{fig:qwen_over_time}
\end{figure}

Across successive Qwen model generations, while parameter size is kept relatively consistent, models become better at identifying which representation is inconsistent once they have detected a disagreement (Figure~\ref{fig:qwen_over_time}). However, the false-positive disagreement rate on clean examples remains high across model generations. While newer models are better at resolving a detected conflict, they are not necessarily better at determining whether a conflict exists in the first place. 

\subsection{Per-model breakdowns}
\label{app:xr-per-model-breakdown}
\begin{figure}[H]
    \centering
    \includegraphics[width=\linewidth]{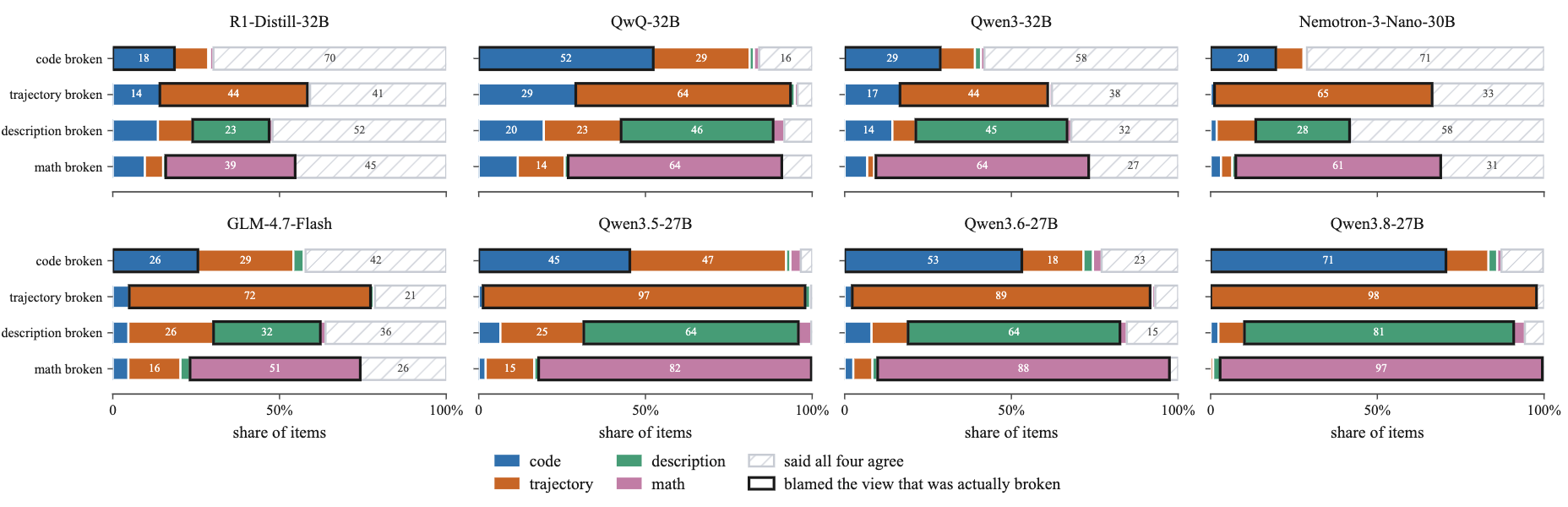}
    \caption{Localization of the inconsistent representation after a disagreement by corruption type, broken down by model. For each corrupted view, bars show the fraction of representations blamed as the outlier, including cases models fail to identify disagreement.}
    \label{fig:app_bar_by_model}
\end{figure}

\begin{figure}[H]
    \centering
    \includegraphics[width=0.8\linewidth]{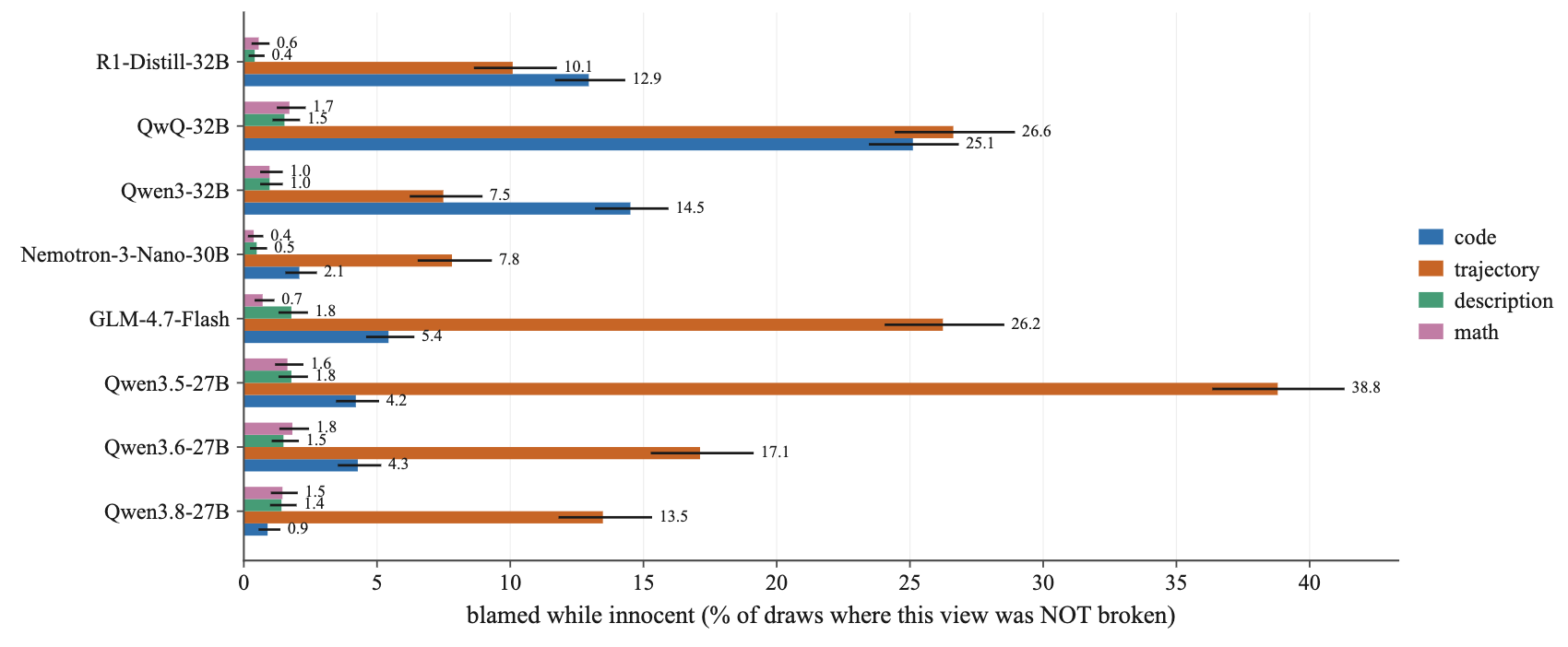}
    \caption{How often each view is blamed on the draws where it was not the corrupted one, broken down by model. Trajectory takes the most innocent blame in 6 of 8 models.}
    \label{fig:blame_by_model}
\end{figure}

\begin{figure}[h]
    \centering
    \includegraphics[width=\linewidth]{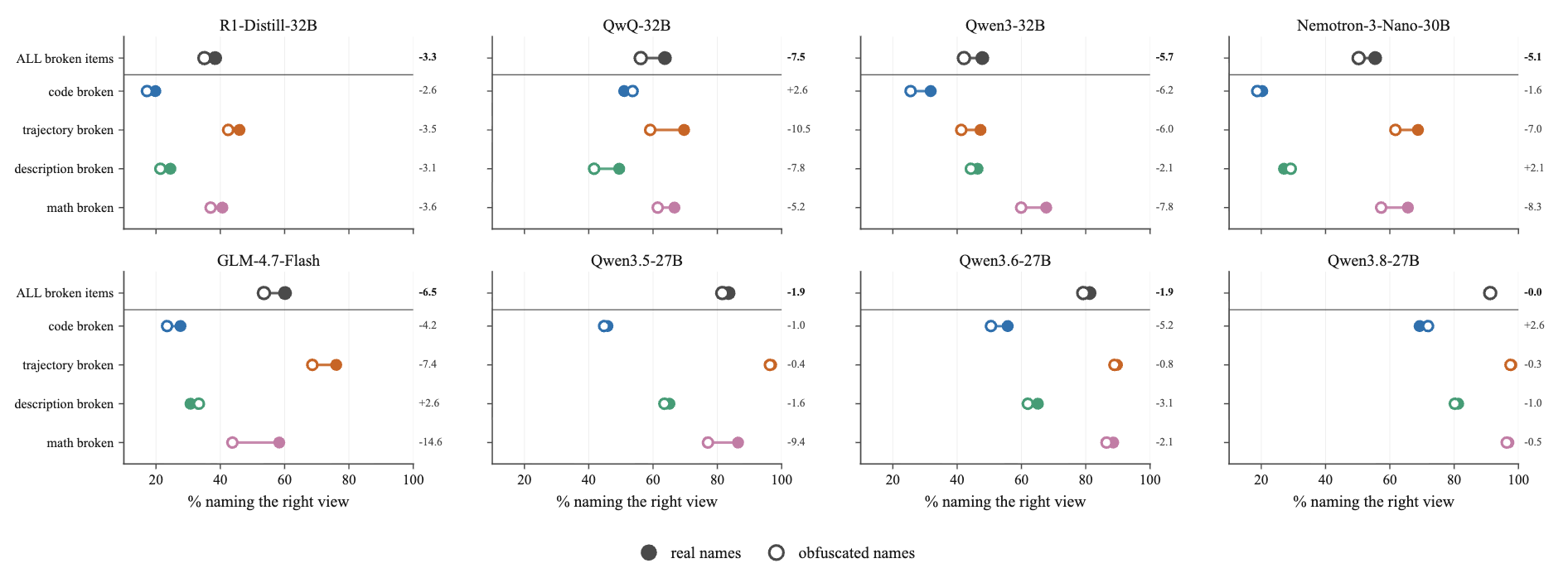}
    \caption{Effect of source-code identifier obfuscation on cross-representation judgments, broken down by model.}
    \label{fig:obf_vars_by_model}
\end{figure}

We disaggregate the cross-representation results by model below.
Figure~\ref{fig:app_bar_by_model} shows inconsistency localization across
corruption types for each model. Figure~\ref{fig:blame_by_model} shows
innocent-blame rates by representation, demonstrating that the tendency
to distrust trajectories is present in six of eight models. Figure~\ref{fig:obf_vars_by_model} shows the effect
of source-code identifier obfuscation separately for each model.

\section{Static validity judgments under lexical perturbations}
\label{app:validity-judgments}

For each solver, we evaluate both the valid and corresponding invalid implementations under four lexical conditions: original comments and identifiers, comments removed, corrupted comments, and obfuscated identifiers. We use the following sampling parameters for all models: $T = 0.6$, $p = 0.95$, top-$k = 20$. For each prompt, we sample $k=3$ generations per model, and prompt the model: \texttt{"Does this simulation produce a physically correct solution for the PDE? Running without error is not sufficient."} All models have reasoning enabled, and generation budgets were set per model to
the largest the checkpoint supports.  

\begin{figure}[h]
    \centering
    \includegraphics[width=0.85\linewidth]{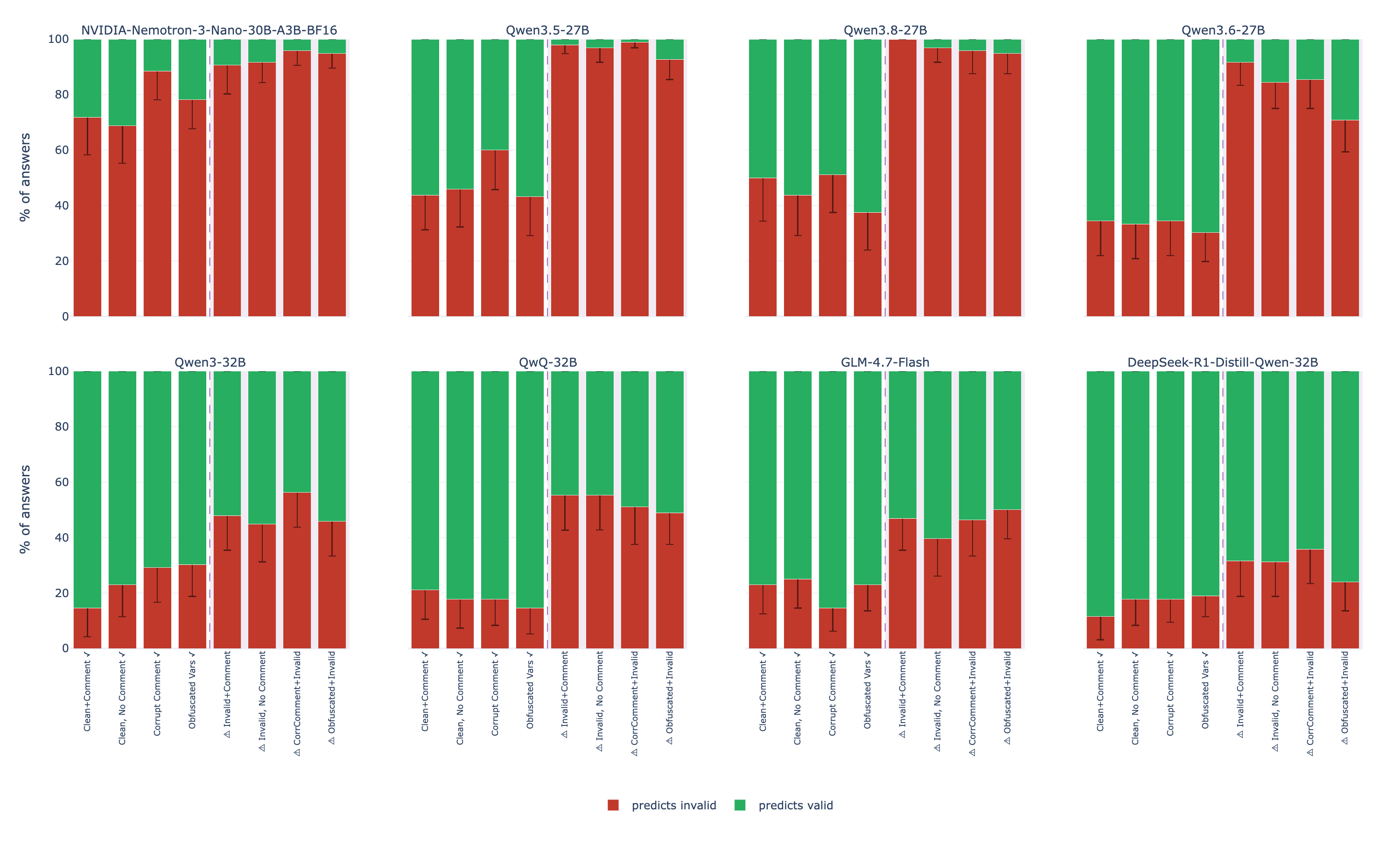}
    \caption{Free-generation validity judgments across lexical conditions, broken down by model. The first four conditions contain valid implementations and the last four contain invalid implementations, separated by the dashed line. Therefore, correct predictions are green on the left and red on the right. Sensitivity to lexical cues varies substantially across models.}
    \label{fig:static-validity-judgments}
\end{figure}

Figure~\ref{fig:static-validity-judgments} shows that validity judgments vary substantially both across models and across perturbations. Models generally struggle on this task and, in several models, removing or corrupting comments or obfuscating identifiers changes the fraction of implementations judged valid.

\newpage

\section{Agentic implementation details}
\label{app:agentic}

\subsection{Models and inference configuration}
\label{app:agent_models}

\begin{table}[h]
\centering
\small
\begin{tabular}{@{}llll@{}}
\toprule
Arm & Model Version & Thinking Control & Setting \\
\midrule
2.5-flash-tb0    & \texttt{gemini-2.5-flash} & \texttt{thinking\_budget} & $0$ (disabled) \\
2.5-flash-tb8192 & \texttt{gemini-2.5-flash} & \texttt{thinking\_budget} & $8{,}192$ tokens \\
3.7-flash-low    & \texttt{gemini-3.7-flash} & \texttt{thinking\_level}  & \texttt{LOW}  \\
3.7-flash-high   & \texttt{gemini-3.7-flash} & \texttt{thinking\_level}  & \texttt{HIGH} \\
\bottomrule
\end{tabular}
\caption{Models and thinking configurations used in the agentic experiments. The
two families expose thinking through incompatible parameters: Gemini 2.5 takes a
numeric \texttt{thinking\_budget} and can disable thinking entirely, whereas
Gemini 3.7 takes a \texttt{thinking\_level}.}
\label{tab:ag_models}
\end{table}

Model versions and configurations are shown in Table~\ref{tab:ag_models}. Each of the four arms is run on all $256$ scripts of the dataset ($32$ ground-truth
samples $\times$ $8$ lexical perturbation conditions), for $1{,}024$ episodes in
total. We sample $k=1$ generation per prompt.

We set $T = 0$ for the Gemini 2.5 arms. For the Gemini 3.7 arms we omit
\texttt{temperature}, \texttt{top\_p}, and \texttt{top\_k} entirely as those arguments are no longer supported. \texttt{max\_output\_tokens} is $50{,}000$ per call in both stages.

\subsection{Prompts}
\label{app:agent_prompt}

Each episode presents the same solver script twice. Stage~1 (S1) is a single static
read with no tools; Stage~2 (S2) continues the same conversation, so the model sees
its own S1 answer in context. The S1 prompt is identical across all
four arms, and both stages share one definition of the validity score, so that an
S1-vs-S2 comparison measures the same question at both stages.

\paragraph{Stage 1.}
\texttt{\{code\}} is substituted with the raw Python source of the solver script:

\begin{promptbox}{Stage 1 prompt}
\begin{Verbatim}[fontsize=\small]
You are analyzing a numerical simulation written in Python.

{code}

Answer the following about this simulation. Be concise.

Output only:
pde: ____
method: ____
behavior: ____
valid: ____
validity_score: ____
pde_exp: ____
method_exp: ____
behavior_exp: ____
valid_exp: ____

- pde: the type of PDE being solved
- method: numerical method(s) used — list all that apply
- behavior: dominant physical process(es) — list all that apply
- valid: does this code run and produce a correct physical solution for the
  PDE?
- validity_score: an integer from 0 to 10 rating physical validity for the
  code - how closely the trajectory from the code would match the intended
  physical system.
0 means it retains none of the intended system's physics; 10 means it
reproduces the intended system's physics exactly.
This is a score of the degree of physical validity, not a confidence score
in the binary valid answer field.
- pde_exp/method_exp/behavior_exp/valid_exp: 1-2 sentence explanation for
  each answer above.
valid_exp should justify both the valid answer and the validity_score — say
why that degree of physical validity, not just yes or no.
\end{Verbatim}
\end{promptbox}

\paragraph{Stage 2.}
Appended as a new user turn after the model's Stage~1 answer. The tool list and
the investigative budget (here $6$) are substituted at build time.

\begin{promptbox}{Stage 2 prompt}
\begin{Verbatim}[fontsize=\small]
In the previous stage, you answered questions about the code snippet's PDE
type, numerical method, behavior, and physical validity just by reading. The
PDE code snippet provided in the previous stage runs without execution
errors but could be logically or physically invalid. You can therefore use
execution to check and, if needed, correct your answers from the previous
stage. The goal of this stage is to find the correct answers about the
original code snippet's PDE type, numerical method, behavior, and physical
validity. You can now use the following tools to collect evidence before
deciding whether to preserve or change answers from the previous stage:
edit_source, run_diagnostic, submit_final_answer, with an investigative
budget of 6 edit_source/run_diagnostic actions total (submit_final_answer
does not count against this budget). Once the budget is used up, only
submit_final_answer remains available. Use the submit_final_answer tool to
provide a structured final answer, do not give a text-only answer. Answers
with both text and an action are preferred. Think about what actions to take
before taking an action such as writing code. Ensure any conclusions based
on execution evidence use specific findings. Text written to stdout and
stderr and saved numeric data (npz, txt, csv) persist between turns, but
images cannot be accessed. Any single file you write must stay under 1GB --
writing more will terminate that turn's execution and delete the file. Avoid
long reasoning written out inside comments in the code. Answers that do not
include tool calls still count against the 6-turn budget, and a text-only
answer will trigger a subsequent turn where you are forced to take an
action.
\end{Verbatim}
\end{promptbox}

\paragraph{Agent tools.}
During Stage~2, the agent has access to three tools summarized in
Table~\ref{tab:agent-tools}. The first two consume one of the six investigative
actions, while \texttt{submit\_final\_answer} does not. The agent may modify and
rerun the simulator to collect new evidence, or analyze previously saved numeric
outputs and source files. Final answers are always evaluated with respect to the
original, unmodified solver.

\begin{table}[h]
\centering
\small
\begin{tabular}{p{0.20\linewidth} p{0.27\linewidth} p{0.40\linewidth}}
\toprule
\textbf{Tool} & \textbf{Purpose} & \textbf{Behavior} \\
\midrule

\texttt{edit\_source}
&
Modify and execute the simulator to collect new evidence.
&
The agent provides a complete replacement source file, which is executed in the
episode sandbox. Printed output is returned to the model, and numeric data may be
saved to disk for later analysis. An empty call reruns the latest version of the simulator.
Consumes one investigative action.
\\

\texttt{run\_diagnostic}
&
Analyze source code or evidence collected from previous executions.
&
Runs a standalone Python diagnostic provided by the agent without rerunning the simulator. It may read
the current source files and numeric files previously saved by
\texttt{edit\_source}. Consumes one investigative action.
\\

\texttt{submit\_final\_answer}
&
Submit the revised PDE, method, behavior, and physical-validity judgment.
&
The answer must concern the original solver and may cite collected execution
evidence. The model reports a binary validity judgment, a 0--10 physical-validity
score, and brief explanations. Does not consume the investigative budget.
\\

\bottomrule
\end{tabular}
\caption{Tools available to the agent during Stage~2.}
\label{tab:agent-tools}
\end{table}

\paragraph{Submission confirmation.}

Before finalization, the harness re-shows the original solver and the model’s Stage~1 answer to prevent confusion between the original and subsequently edited code. This occurs both when the investigative budget is exhausted and when the agent voluntarily calls \texttt{submit\_final\_answer}; in the latter case, the first submission is intercepted and a second is required. Thus, early and forced submissions answer from equivalent context. This was introduced after we observed self-inflicted
evidence, in which a model reported a defect it had itself introduced while
rewriting, as a failure mode during preliminary experiments.

\subsection{Agent design}
\label{app:agent_design}

The agentic experiments used the hosted Gemini API.

An episode is one agentic run over a single code snippet. It begins when the
model is shown the code and its own Stage-1 judgment, and ends when the final
Stage-2 answer is submitted. Within an episode the model takes turns; a turn is one model generation together
with the environment's response to it. The model is given three tools--predefined functions, detailed in Table~\ref{tab:agent-tools}.

We budget how many turns the model may spend investigating. An action is a turn
that consumes this budget. A turn invoking \texttt{edit\_source} or
\texttt{run\_diagnostic} consumes one of six actions. The model is instructed to
return a tool call and optionally text on every turn, but occasionally returns no tool call through omission or a malformed call. Such a turn also consumes an action, and the model is then
forced to emit only a tool call on the following turn. Because
\texttt{submit\_final\_answer} is not investigative, we define it as a tool call
but not an action, and it consumes no budget. An episode may therefore contain
more turns than actions.

\paragraph{Environment and sandboxing.}
Each episode executes in its own working directory, and each subprocess is
capped at 120 seconds. Writes are bounded by a $1$\,GB per-file hard cap,
which prevents degenerate behaviour such as serializing an entire trajectory,
and a 1.5\,GB cumulative cap on the episode directory. Absolute paths are
scrubbed from \texttt{stdout} and \texttt{stderr}, and the episode directory is
named by a SHA-256 digest of the snippet title, so that neither the directory
name nor the captured output exposes the ground-truth label. A directory snapshot is saved after every turn for post-hoc inspection.

\paragraph{Harness.} The harness is written for this task and broadly follows the Antigravity design.
The investigative budget is 6 actions. \texttt{edit\_source} takes full-file
rewrites rather than diffs or patches: the snippets are short, and Gemini~2.5
produced malformed patches more often than malformed full files. Tool results
are truncated to 4000 characters and generation is capped at 50000 output tokens per call. A text-only turn is followed by a turn on which tool use
is forced. The prompt discourages calling \texttt{run\_diagnostic} before any
evidence has been saved and discourages writing image files, and asks for a tool
call on every turn. An episode aborts after two protocol violations.

\subsection{Additional validity results}
\label{app:additional-validity-agents-results}

\paragraph{Validity judgment as a discrimination and calibration problem.} We note that assessing physical validity can be divided into a discrimination problem (whether the model can find issues in the simulation) and a calibration problem (whether those issues warrant an invalid label). We also note that our dataset is not high-fidelity simulation code -- we collect textbook-style illustrative implementations of well-characterized equations that produce qualitatively physically correct behavior. However, LLMs may still critique implementation choices in these solvers that affect simulation quality. As such, we observe that models are much better at finding defects than discriminating and calibrating their results (as seen in Figure \ref{fig:agentic2}). To ensure that our results are not confounded by different thresholding between LLMs and human dataset authors, we ask the model to produce both binary validity judgments and a score from 0-10. 

\begin{figure}[h]
    \centering
    \includegraphics[width=.9\linewidth]{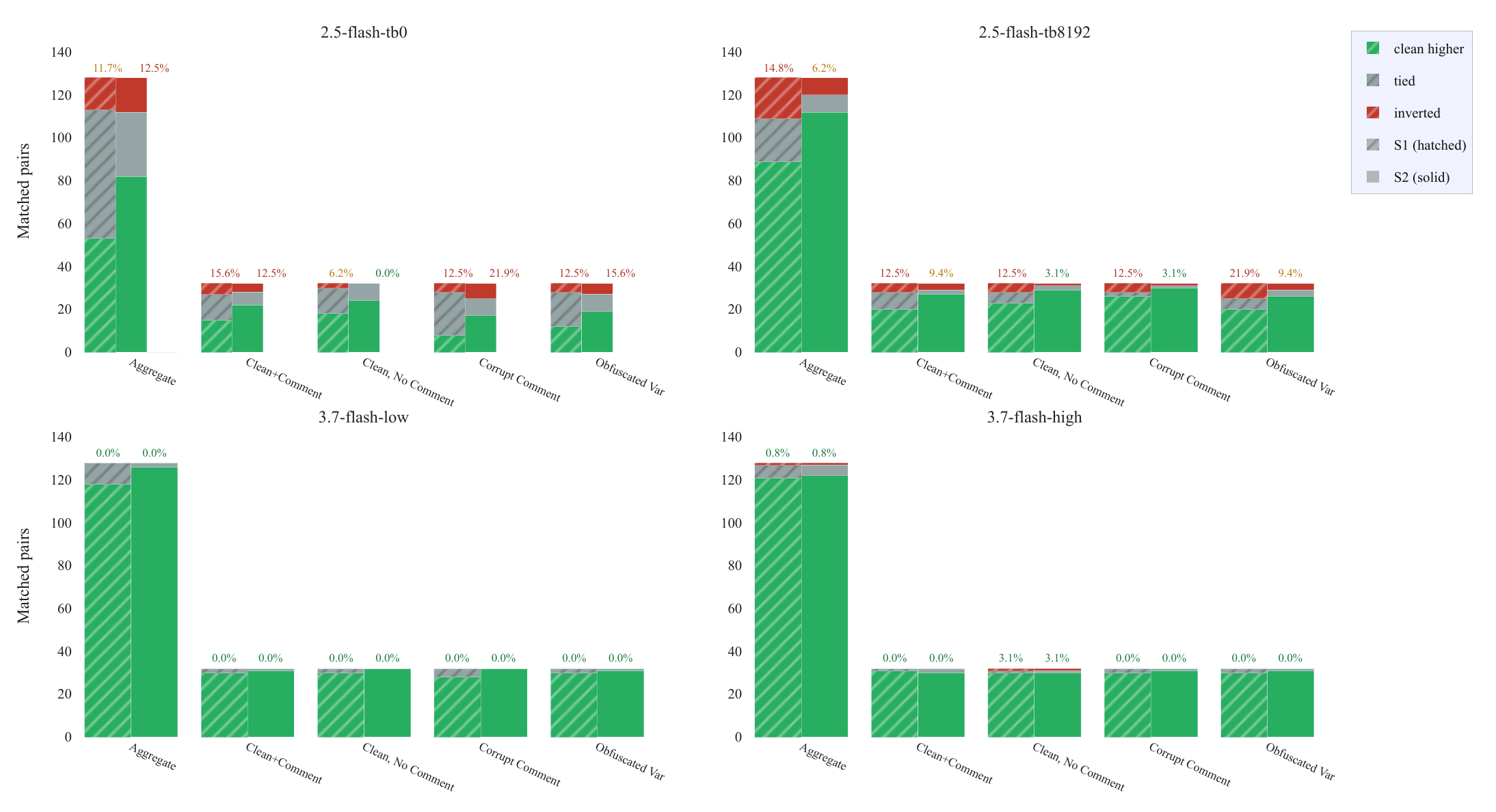}
    \caption{Paired valid-invalid rankings by each agent at S1 (static code reading) vs S2 (agentic execution) across code perturbations.}
    \label{fig:agenticsuppfig1}
\end{figure}

\paragraph{Analysis of inversions.} To distinguish when models simply cannot calibrate validity scores versus when they genuinely cannot discriminate, we measure \emph{inversions}, or cases when a model scores an invalid solver better on validity than its valid counterpart. Figure~\ref{fig:agenticsuppfig1} compares the relative validity scores assigned
to matched valid--invalid solver pairs before and after execution. Execution does not help Gemini 2.5 without thinking
(12\% to 13\% inversions), but reduces inversions with thinking
(15\% to 6\%). Gemini 3.7 is considerably more robust, with inversion rates at
or near zero in both stages. The effect also varies across lexical conditions,
with the largest changes for Gemini 2.5.

\paragraph{Qualitative failure cases.} 
\label{qual-failure-case} 
\begin{enumerate}
    \item  \textbf{Gemini 3.7, high thinking, Wave equation episode:} 
First, we show an example from the best agent, where execution does not always lead to physically meaningful verification. In one Wave equation episode, Gemini 3.7 incorrectly labeled an invalid solver valid. The reason was that it failed to 
identify a spurious forcing term that violated physical conservation laws, but concluded that saved execution values matched expected values within tolerance. The agent therefore
performed an apparently quantitative verification while testing behavior
produced by an already incorrect simulator. This illustrates how execution
can reinforce an incorrect physical interpretation when the diagnostic itself
does not target the causal defect.
\item

\textbf{Gemini 2.5, no thinking, Burgers' equation episode:}
Next, we show an example where a weaker agent (Gemini 2.5, no thinking) correctly identifies an invalid solver but fails to identify the causal defect. In one invalid Burgers' example with a sign flip, the agent's validity justification states:

\begin{Verbatim}[breaklines=true,
breakanywhere=true,  breaksymbolleft={},
breaksymbolright={},
fontsize=\small]
The simulation is numerically unstable, as evidenced by the presence of NaN values in the solution array `u` after a few time steps. This is likely due to an incorrect calculation of the shock speed `s` at the cell interfaces, leading to an incorrect upwind flux determination. The assertion `assert f_interface.shape[0] == u.shape[0] + 1` passes, but the code still produces an invalid solution.
\end{Verbatim}

While the agent rewrites the solver to save trajectories and then runs diagnostics, the sign flip is completely missed. This is identified by the judge:

\begin{Verbatim}[breaklines=true,
breakanywhere=true,  breaksymbolleft={},
breaksymbolright={},
fontsize=\small]
The model incorrectly claims that the shock speed `s` is calculated incorrectly, but this calculation is identical to the base code. It completely misses the actual discrepancy, which is the reversed sign in the calculation of `du_dt` (subtracting `f_interface[0:n]` from `f_interface[1:n+1]` instead of the other way around).
\end{Verbatim}

In contrast, Gemini 3.7 at high thinking correctly identifies the sign flip in the calculation of $\frac{du}{dt}$ with only static reading. The model then edits the solver with small checks, and cites the exact step and time at which the simulation blows up to infinity in its final justification.

\end{enumerate}

\subsection{Reasoning quality evaluation with LLM-as-judge}
\label{llm-as-judge}

\textbf{Results.} Binary validity accuracy rewards a correct verdict regardless of reasoning, so we additionally judge \emph{why} a model called code invalid. We score only rows
where the model correctly labels the snippet as invalid after S2
($\text{gt\_valid}=\text{False}$ and the stage's verdict is correct): $121$, $120$,
$123$ and $124$ rows for the four arms respectively, $488$ in total. Correct
verdicts on valid code have no defect to name and are excluded.

The judge is \texttt{gemini-3.1-pro-preview} \citep{gemini31pro}, which we chose as it is a separate, stronger model than either Flash model used in the agent. It receives the known-valid base version of the same simulation, the perturbed version the model evaluated, a human annotator's note on the expected symptom, and the model's own justification; it does not receive the model's execution output. Function calling is forced to a single \texttt{submit\_judgment} tool. The judge sees code whose comments are absent, obfuscated, or deliberately
misleading. For the three conditions where this would impede judgment, a
genuine-comment reference version is appended after each code block, preceded by
a note stating what was altered. Because $89\%$ of eligible justifications state
the validity score in prose, blinding is unavailable; the prompt instead
instructs the judge to disregard it. The full prompt is given below.

\textbf{Human audit of judge results.} We assess the reliability of the LLM-as-a-judge evaluation by conducting a manual audit of 20 cases where an agent correctly classified a corrupted PDE solver as invalid. We randomly sampled 20 justifications across both models (5 per thinking setting). A human expert was given access to these 20 cases and provided a rubric to judge if the agent's justification was correct. The expert filled out the same fields as the LLM judge, and did not have access to the LLM judge's annotations. Finally, we compared the assessments by the human expert and the LLM judge. Both have high agreement of at least 90\% for category and incorrect code claims (Table \ref{tab:judge-audit}). Both incorrect-claim disagreements are Gemini 2.5 Flash items flagged by the LLM judge but not by the expert (LLM judge 6/10 vs expert 4/10) (Table \ref{tab:judge-audit-confusion}). This suggests that the LLM judge's 2.5 incorrect-claim rate may be slightly overestimated. Overall, the human audit supports the reliability of the LLM-as-a-judge evaluation in distinguishing correct causal explanations from incomplete or partially incorrect reasoning.

  \begin{table}[H]
  \centering
  \small
  \begin{tabular}{p{0.28\linewidth} p{0.25\linewidth} p{0.14\linewidth} p{0.14\linewidth}}
  \toprule
  Measure & Pooled ($n=20$) & Gemini 2.5 Flash ($n=10$) & Gemini 3.7 Flash ($n=10$) \\
  \midrule
  Exact category (none / some / all) & 18/20 (90\%) [70--97\%] & 9/10 & 9/10 \\
  Named $\geq$1 real defect & 19/20 (95\%) [76--99\%] & 9/10 & 10/10 \\
  Incorrect code claims & 18/20 (90\%) [70--97\%] & 8/10 & 10/10 \\
  \bottomrule
  \end{tabular}
  \caption{Agreement between a PDE expert and the LLM judge on 20 examples. Cells show the fraction of items where the expert and the judge agree, the percentage and a 95\% Wilson interval in the second column. ``Named $\geq$1 real defect'' indicates ``some'' or ``all.'' }
  \label{tab:judge-audit}
  \end{table}

\begin{table}[H]
\centering
\small
\begin{tabular}{>{\raggedright\arraybackslash}p{0.14\linewidth}
                >{\raggedright\arraybackslash}p{0.17\linewidth}
                >{\raggedright\arraybackslash}p{0.17\linewidth}
                >{\raggedright\arraybackslash}p{0.17\linewidth}
                >{\raggedright\arraybackslash}p{0.17\linewidth}}
\toprule
 & \multicolumn{2}{c}{Named $\geq$1 real defect} & \multicolumn{2}{c}{False code claims} \\
\cmidrule(lr){2-3} \cmidrule(lr){4-5}
Expert & Judge: yes & Judge: no & Judge: yes & Judge: no \\
\midrule
yes & 16 (6 / 10) & 1 (1 / 0) & 4 (4 / 0) & 0 \\
no  & 0 & 3 (3 / 0) & 2 (2 / 0) & 14 (4 / 10) \\
\bottomrule
\end{tabular}
\caption{PDE expert (rows) vs.\ LLM judge (columns) on the same 20 audited justifications. Diagonal cells are agreement. Counts are pooled across models and the Gemini 2.5 Flash /
Gemini 3.7 Flash split is in parentheses.}
\label{tab:judge-audit-confusion}
\end{table}
  
\textbf{Prompt for LLM-as-judge reasoning evaluation.}

\begin{promptbox}{LLM-as-judge prompt}
\begin{Verbatim}[fontsize=\small]
You are evaluating whether a model's stated justification for why a piece of
PDE solver code is physically/numerically invalid actually corresponds to a
real discrepancy in the code, compared to a known-valid reference version of
the same simulation.

{base_valid_code}

{corrupted_code}

For context, here is a human annotator's note on the general symptom this
corruption is expected to cause (this is a general description, not
necessarily a specific mechanism -- do not treat it as the definitive or
only correct answer):

{invalidity_note}

The model was asked to give a binary answer and a score for validity. Every
example provided here was one the model answered "invalid" for the binary
field. Its justification may mention the assigned validity score. The task
here is to evaluate the accuracy of the claims in valid_exp solely with
respect to the actual invalidities in the code, not the model's stated
validity score.

The model was able to execute the code before giving its answer, so claims
about runtime behaviour may be genuine observations rather than speculation.
You cannot verify them -- you have the code, not the execution output. Judge
claims about the code on their correctness; record unverifiable runtime
figures in the separate field rather than counting them as errors.

Here is the model's own justification for why it judged the corrupted code
invalid:

{model_justification}

Compare the base and corrupted code (and reference versions, if provided)
directly to determine what actually changed. Then call submit_judgment with
your assessment.
\end{Verbatim}
\end{promptbox}

\paragraph{Judge outputs.}
The judge outputs (1) whether the justification identifies none, some, or all genuine defects, (2) whether it makes any incorrect claims about the code, and (3) whether it cites execution-specific evidence that cannot be verified from static inspection. The judge also provides a brief rationale for these labels.



\end{document}

%% file: figure_tikz_arxiv.tex
\providecommand{\pdeicon}[1]{\tikz[baseline=-0.55ex,x=1mm,y=1mm,line width=0.35pt,rounded corners=0pt,line join=miter,>=stealth]{#1}}
\providecommand{\iconmag}{\pdeicon{\draw (0,0.2) circle (1.05); \draw[line width=0.45pt] (0.78,-0.58)--(1.75,-1.55);}}
\providecommand{\iconrun}{\pdeicon{\fill (-0.95,-1.15)--(-0.95,1.15)--(1.25,0)--cycle;}}
\providecommand{\iconrev}{\pdeicon{\draw[line width=0.4pt] ([shift={(0,0.35)}]0:0.95) arc (0:-296:0.95); \fill[shift={(0,0.35)}, shift=(64:0.95), rotate=-26] (0.42,0)--(-0.28,0.30)--(-0.14,0)--(-0.28,-0.30)--cycle;}}
\providecommand{\iconok}{\pdeicon{\draw[green!55!black,line width=0.5pt] (-1.05,0.1)--(-0.35,-0.75)--(1.1,1.0);}}
\providecommand{\iconbad}{\pdeicon{\draw[red!75!black,line width=0.5pt] (-0.9,-0.9)--(0.9,0.9) (-0.9,0.9)--(0.9,-0.9);}}
\providecommand{\icondb}{\pdeicon{\draw (0,0.62) ellipse (1.05 and 0.4); \draw (-1.05,0.62)--(-1.05,-0.62); \draw (1.05,0.62)--(1.05,-0.62); \draw (-1.05,0) arc (180:360:1.05 and 0.4); \draw (-1.05,-0.62) arc (180:360:1.05 and 0.4);}}
\providecommand{\iconspark}{\pdeicon{\draw[line width=0.3pt] (-2.1,-1.3)--(-2.1,1.5) (-2.1,-1.3)--(2.3,-1.3); \draw[line width=0.5pt] (-1.8,-0.7)--(-0.7,0.9)--(0.4,-0.2)--(1.4,1.2)--(2.1,0.4);}}
\providecommand{\iconquote}{\pdeicon{\draw[rounded corners=0.5] (-2.2,-0.5) rectangle (2.2,1.9); \draw (-1.2,-0.5)--(-1.5,-1.6)--(-0.35,-0.5); \draw[line width=0.45pt] (-1.5,1.25)--(1.5,1.25) (-1.5,0.45)--(0.7,0.45);}}
\begin{figure*}[t]
\centering
\resizebox{\textwidth}{!}{%
\begin{tikzpicture}[
    >=stealth,
    y=1.30cm,
    font=\footnotesize,
    box/.style={draw, thick, rounded corners=3pt, fill=white, align=center, inner sep=3pt},
    view/.style={box, text width=2.20cm, minimum height=0.90cm, inner xsep=1pt, font=\footnotesize},
    flowbox/.style={font=\footnotesize, align=center, text width=1.55cm, inner sep=2pt},
    pbox/.style={draw, thick, rounded corners=4pt}
]

\node[font=\bfseries\normalsize, text=blue!80!black] (t1_title) at (-5.00, -0.82)
    {Task 1: Cross-Representation Consistency};

\node[view] (v1) at (-8.51, -1.72) {\textbf{Code} $\boldsymbol{(C_s)}$\\[2pt]\texttt{</>}};
\node[view] (v2) at (-6.17, -1.72) {\textbf{Math} $\boldsymbol{(M_s)}$\\[2pt]$\boldsymbol{\partial_t u = \dots}$};
\node[view] (v3) at (-3.83, -1.72) {\textbf{Trajectory} $\boldsymbol{(T_s)}$\\[2pt]\iconspark};
\node[view] (v4) at (-1.49, -1.72) {\textbf{Description} $\boldsymbol{(D_s)}$\\[2pt]\iconquote};

\draw[dashed, rounded corners=3pt] (-9.68, -1.22) rectangle (-0.32, -2.42);
\node[fill=blue!5, font=\footnotesize, inner xsep=4pt, inner ysep=2pt] at (-5.00, -2.42)
    {Four views of the same physical system $\boldsymbol{s}$};

\node[font=\footnotesize] (t1_text) at (-5.00, -2.78) {
    \textcolor{green!55!black}{\textbf{\checkmark} \textbf{Consistent:}} all four views describe $\boldsymbol{s}$
    \quad
    \textcolor{red!75!black}{$\boldsymbol{\times}$ \textbf{Inconsistent:}} one view disagrees
};

\node[box, font=\footnotesize, inner sep=3pt] (t1_lex) at (-5.00, -3.22) {
    \textbf{Code identifiers:} informative $\boldsymbol{\longleftrightarrow}$ obfuscated
};

\node[box, fill=blue!15, draw=blue!50!black, font=\footnotesize, inner sep=3pt, minimum width=9.20cm]
    (t1_q) at (-5.00, -3.72)
    {\textbf{LLM is asked:} do the four views agree? If not, which view is the outlier?};

\begin{scope}[on background layer]
\draw[pbox, fill=blue!5, draw=blue!40!black] (-9.80, -0.50) rectangle (-0.20, -4.10);
\end{scope}

\node[font=\bfseries\normalsize, text=green!50!black] (t2_title) at (5.00, -0.82)
    {Task 2: Static and Agentic Validity};

\node[box, text width=2.60cm, minimum height=0.90cm, font=\footnotesize] (valid) at (2.25, -1.72)
    {\iconok~\textbf{Valid Solver}\\[2pt]\texttt{\textbf{+nu}*laplacian(u)}};
\node[box, text width=2.60cm, minimum height=0.90cm, font=\footnotesize] (invalid) at (7.75, -1.72)
    {\iconbad~\textbf{Invalid Solver}\\[2pt]\texttt{\textcolor{red!75!black}{\textbf{-nu}}*laplacian(u)}};
\draw[->, thick] (valid) -- node[above, font=\footnotesize\bfseries, inner sep=2pt] {Physical defect} (invalid);

\node[box, font=\footnotesize, inner sep=3pt, minimum width=9.20cm] (t2_lex) at (5.00, -2.78) {
    \textbf{Lexical variants:} original, no comments, corrupted comments, obfuscated IDs
};

\node[box, minimum width=9.20cm, minimum height=0.90cm] (agentbar) at (5.00, -3.50) {};
\node[font=\footnotesize\bfseries, anchor=west] (agentlab) at (0.60, -3.50) {LLM agent:};
\node[flowbox] (s1) at (3.27, -3.50) {\iconmag~Static\\judgment};
\node[flowbox] (s2) at (5.80, -3.50) {\iconrun~Execute \&\\diagnose};
\node[flowbox] (s3) at (8.33, -3.50) {\iconrev~Revised\\judgment};

\draw[->, thick] (s1) -- (s2);
\draw[->, thick] (s2) -- (s3);

\begin{scope}[on background layer]
\draw[pbox, fill=green!5, draw=green!40!black] (0.20, -0.50) rectangle (9.80, -4.10);
\end{scope}

\end{tikzpicture}%
}
\vspace{-8pt}
\caption{Experimental overview of Solver2System's two evaluation settings.}
\label{fig:overview}
\vspace{-4pt}
\end{figure*}